\documentclass{article} 
\usepackage{iclr2026_conference,times}
\iclrfinalcopy

\usepackage{amsmath,amsfonts,bm}

\def\eqref#1{equation~\ref{#1}}

\def\1{\bm{1}}

\DeclareMathAlphabet{\mathsfit}{\encodingdefault}{\sfdefault}{m}{sl}
\SetMathAlphabet{\mathsfit}{bold}{\encodingdefault}{\sfdefault}{bx}{n}

\usepackage{hyperref}
\hypersetup{colorlinks=true, citecolor=blue, linkcolor=blue, urlcolor=blue}
\usepackage{url}

\usepackage{amsmath}
\usepackage{graphicx}
\usepackage{multirow}
\usepackage{multicol}
\usepackage{float}
\usepackage{caption}
\usepackage{wrapfig}
\usepackage[capitalise]{cleveref}
\usepackage{subcaption}
\usepackage{xspace}
\usepackage{comment}
\usepackage{booktabs}
\usepackage{algorithm}
\usepackage{algpseudocode}

\usepackage[most]{tcolorbox}
\usepackage{lipsum}

\title{Efficiently Estimating Data Efficiency for Language Model Fine-tuning}

\author{%
  Gyung Hyun Je \\
  University of Toronto \\
  \texttt{jayje@cs.toronto.edu} \\
  \And
  Colin Raffel \\
  University of Toronto\\
  \texttt{craffel@gmail.com} \\
}

\newcommand{\NameOfMeasurement}{\texttt{CoS-Low}\xspace}

\begin{document}
\maketitle

\begin{abstract}

While large language models (LLMs) demonstrate reasonable zero-shot capability across many downstream tasks, fine-tuning is a common practice to improve their performance. However, a task's \textit{data efficiency} --- i.e., the number of fine-tuning examples needed to achieve a desired level of performance --- is often unknown, resulting in costly cycles of incremental annotation and retraining. Indeed, we demonstrate across a curated set of 30 specialized tasks that performant LLMs may struggle zero-shot but can attain stronger performance after fine-tuning. This motivates the need for methods to predict a task's data efficiency \textit{without} requiring incremental annotation. After introducing a concrete metric that quantifies a task's data efficiency, we propose using the \textit{gradient cosine similarity of low-confidence examples} to predict data efficiency based on a small number of labeled samples. We validate our approach on a diverse set of tasks with varying data efficiencies, attaining 8.6\% error in overall data efficiency prediction and typically eliminating hundreds of unnecessary annotations on each task. Our experiment results and implementation code are available on GitHub \footnote{\url{https://github.com/r-three/dataefficiency}}.

\end{abstract}

\section{Introduction} \label{sec:introduction}

Large language models (LLMs) are increasingly treated as generalist systems that can competently perform any text-based task zero-shot, i.e.,\ without requiring any task-specific training data \citep{brown2020languagemodelsfewshotlearners}.
However, the zero-shot performance of an LLM often lags behind human-level (or otherwise acceptable) performance  (\cite{li2023chatdoctormedicalchatmodel,liu2022fewshotparameterefficientfinetuningbetter, ouyang2022traininglanguagemodelsfollow,sanh2022multitaskpromptedtrainingenables,singhal2023expertlevelmedicalquestionanswering,wei2022finetunedlanguagemodelszeroshot}).
In such cases, fine-tuning on task-specific data can provide a simple way to improve an LLM's performance by reinforcing the specified format of the model response (\cite{ouyang2022traininglanguagemodelsfollow,sanh2022multitaskpromptedtrainingenables,wei2022finetunedlanguagemodelszeroshot}) or specializing the LLM to the task (\cite{li2023chatdoctormedicalchatmodel,liu2022fewshotparameterefficientfinetuningbetter,singhal2023expertlevelmedicalquestionanswering}).
Indeed, fine-tuning a pre-trained LLM can require orders of magnitude less task-specific data than training on the task from scratch. \cite{zhou2023limaalignment} show that an LLM can easily learn to output high-quality responses with only hundreds or thousands of examples, which \cite{aghajanyan2020intrinsicdimensionalityexplainseffectiveness} suggests is enabled by the pretraining phase compressing large-scale knowledge and reducing the downstream task's intrinsic dimensionality.

A key consideration when fine-tuning LLMs is the task's \textit{data efficiency}, i.e.,\ the number of task-specific labeled data points required to reach a desired performance level.
Unfortunately, the data efficiency of a given task is generally not clear a priori -- as we show in \cref{sec:data-efficiency-variability}, some tasks require only a few dozens of samples to reach or exceed human-level performance while others may require thousands.
A straightforward way of determining a task's data efficiency is to collect a large pool of labeled data and fine-tune the model at various data budgets, evaluating performance at each budget and determining the amount of data required to reach desired performance.
However, this approach requires annotating a large training dataset and fine-tuning many models, obviating the purpose of estimating the data efficiency in the first place.
Fine-tuning scaling laws can be fit to explore the relationship between the model loss and fine-tuning data size \citep{zhang2024scalingmeetsllmfinetuning}, but fitting these scaling laws involves specific parameters unknown before fine-tuning on the downstream task.
We argue that a useful method for predicting fine-tuning data efficiency should be able to do so \textit{efficiently} -- i.e.,\ based on a small number of task-specific labeled examples and requiring a small amount of computation.

\begin{figure}[t]
    \includegraphics[width=\linewidth]{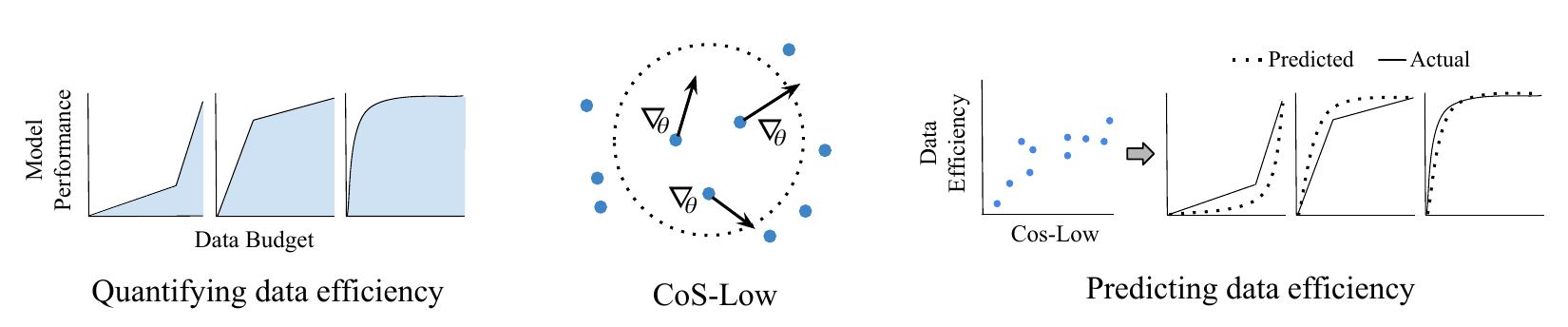}
    \caption{Overview of our approach to predict task-specific data efficiency from a few labeled data points. We formally define a task's ``data efficiency'' as the area under its \textit{data efficiency curve} (left). Then, we show that the aggregate gradient cosine similarity among low-confidence examples (\NameOfMeasurement) is a strong predictor of data efficiency. We propose a method leveraging \NameOfMeasurement to predict the task's data efficiency and its data efficiency curve, providing a concrete estimation of the fine-tuning data size needed to reach a target performance.}
    \label{fig:method-diagram}
\end{figure}

In this work, we propose a method that meets our desiderata for estimating data efficiency (\cref{fig:method-diagram}).
Specifically, we first introduce a precise definition of data efficiency based on the area under the task data efficiency curve.
We then explore different cheaply computable metrics that are predictive of data efficiency and ultimately find that the per-sample gradient cosine similarity of low-confidence examples (\NameOfMeasurement) is highly correlated with our notion of data efficiency, even when computed over a small number of labeled examples.
We then formulate a procedure for estimating data efficiency that maps \NameOfMeasurement to a parametrized approximation of the data efficiency curve.
We validate the effectiveness of this procedure on a curated set of 30 realistic specialized tasks (spanning applications in law, medicine, and well-known benchmarks) with varying levels of data efficiency.
Our approach only requires collecting a small number of labeled examples and does not require fine-tuning or tracking training dynamics, making it a viable option for practitioners in resource-constrained settings that need to determine the number of examples to annotate to reach desired performance on a downstream task.

\section{Establishing the Variability of Data Efficiency}
\label{sec:data-efficiency-variability}

A core assumption in our work is that the data efficiency --- i.e., the relationship between the number of examples used for fine-tuning and performance --- varies significantly from task to task.
To support this assumption, we first curate a diverse set of 30 tasks from multiple domains, including science, medicine, law, finance, sports, customer inquiries, and natural language understanding. These tasks are sourced from popular datasets from HuggingFace as well as well-known benchmarks such as SuperGLUE \citep{wang2020supergluestickierbenchmarkgeneralpurpose}, GLUE \citep{wang2019gluemultitaskbenchmarkanalysis}, and BIG-Bench \citep{srivastava2023imitationgamequantifyingextrapolating}. We mainly consider multi-class text classification and question-answering (QA) to allow consistent use of the exact string match accuracy to measure performance. We also consider the application of our method to generative tasks in \cref{sec:discussion}.
As labeled data and compute resources are limited, finding the true ceiling of model's performance on a given task is challenging. Therefore, we limit our selection to tasks with a known estimate of human-level performance and use this as a proxy of maximum attainable performance for each task. To address cases where human-level performance underestimates this ceiling, we use the higher of the human-level and the maximum performance observed within maximum fine-tuning data budget. Additionally, for method design and evaluation we mainly consider tasks with at least 5000 available labeled examples so that we can measure performance up to a relatively high data budget. We report on tasks with fewer than 5000 labeled examples if the maximum performance is reached after fine-tuning on the available data points. We release the set of prompts used for fine-tuning and evaluation on GitHub, and further details on our chosen tasks are available in  \cref{appendix:downstream-task-details}.

We fine-tune the Llama 3.1 8B Instruct model \citep{grattafiori2024llama3herdmodels} on our set of downstream tasks to evaluate performance after fine-tuning on varying data sizes. Measuring the performance on every possible fine-tuning data size between 1 and 5000 would require 5000 fine-tuning runs for each task and therefore be prohibitively expensive. Instead, we fine-tune the model with 50, 100, 200, 500, 1000, 2500, and 5000 randomly selected data points. We use full model fine-tuning instead of parameter-efficient fine-tuning (PEFT) techniques such as LoRA \citep{hu2021loralowrankadaptationlarge}, as PEFT methods can exhibit a notable performance gap compared to full model fine-tuning (\cite{biderman2024loralearnsforgets,zhang2024scalingmeetsllmfinetuning}) and greater sensitivity to the choice of hyperparameters. Our choice of hyperparameter and training settings are listed in \cref{appendix:fine-tuning-detail}.

Our empirical results, shown in \cref{fig:overall-task-data-efficiency}, demonstrate that the relationship between fine-tuning data size and performance is highly task-dependent.
Some tasks (top rows of figure \cref{fig:overall-task-data-efficiency}) show little to no improvement in the lower data regime but display a substantial boost in performance after a certain inflection point. The others (bottom row of \cref{fig:overall-task-data-efficiency}) show an almost immediate increase in accuracy with as few as 50 fine-tuning examples.
However, across the 30 tasks, performance varies most at smaller fine-tuning data sizes and tends to plateau before 2500 to 5000 examples (\cref{appendix:ablation-data-budget-10k}, \cref{tab:raw-acc-gain}).
Later in \cref{sec:measuring-data-efficiency}, we will define a metric that captures this variability in data efficiency.

Notably, zero-shot accuracy of a task does not necessarily correlate with task data efficiency; tasks with similarly low or high zero-shot accuracies can have widely different task data efficiencies (\cref{appendix:downstream-task-details}).


\section{Measuring Data Efficiency} \label{sec:measuring-data-efficiency}

\begin{figure}
    \includegraphics[width=\linewidth]{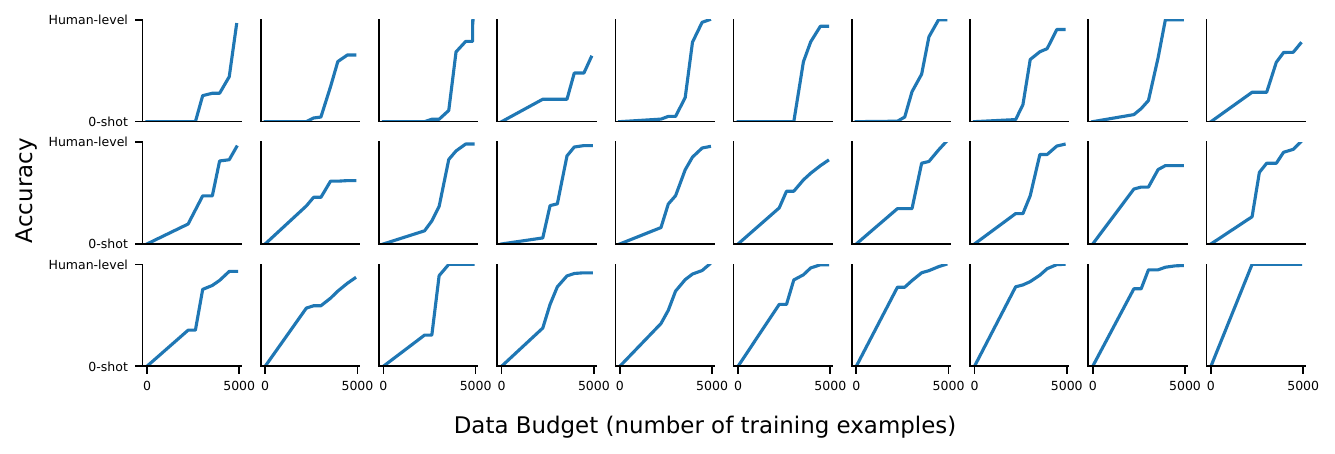}
    \caption{Comparing data budget (from 0 to 5000 examples on log-scale, x-axis) and task performance (from zero-shot to human-level performance, y-axis) across the 30 downstream tasks. The plots are sorted by speed of convergence to the maximum performance level as the fine-tuning data size increases.}
    \label{fig:overall-task-data-efficiency}
\end{figure}

As discussed above, we informally define the ``data efficiency'' of a given task as the extent to which we expect performance to improve when annotating and training on additional training examples.
In other words, the more data efficient a task is, the fewer data points are required for the model to ``solve'' it.
To better quantify this notion, we require a precise measurement that reflects our definition.

To formulate a reliable metric for data efficiency, we make a series of assumptions inspired by the results of \cref{sec:data-efficiency-variability}.
First, we assume that, across all tasks, there is limited benefit in annotating additional examples above some maximum data budget.
In \cref{fig:overall-task-data-efficiency}, performance for many tasks has reached close to the estimated maximum attainable performance or otherwise plateaued by 5000 examples, so we assume that for this choice of base model there is limited room for improvement after annotating 5000 examples.
This assumption will prove valuable later when we aim to map the prediction of our data efficiency metric back to a concrete estimate of the number of training examples required to reach a certain level of performance.
Second, the data efficiency curves assume the model performance at any given data budget is at least as good as the performance at the lower budgets.
Though empirically observed to be true, in the rare cases where this does not hold, we replace the worse performance at the higher budget with the highest observed performance at the lower budgets.
In practice, a practitioner would just choose to train on a smaller data subset if training on \textit{more} data resulted in \textit{worse} performance, which motivates defining a data efficiency curve that captures the best attainable performance achievable \textit{up to} a given data budget.


\subsection{Defining Data Efficiency} \label{sec:data-efficiency-definition}


Having motivated our notion of data efficiency and stated our assumptions, we now introduce our proposed metric for concretely measuring data efficiency. Given a fine-tuning data budget $n \in [0,N]$ where $N=5000$ is the maximum available dataset size, task $k$, and performance function for task $k$ $f_k(n): n \rightarrow \text{acc}_k$ where $\text{acc}_k \in [0,1]$ is a normalized accuracy that maps the raw zero-shot and maximum attainable (human-level) performance to 0 and 1 respectively, we define area under the curve (AUC) of $f_k(n)$ as the data efficiency measure:
$$\text{AUC}_{k} = \frac{1}{N} \sum_{n=0}^{N} f_k(n)$$
where $\text{AUC}_k \in [0,1]$.
Mathematically, our notion of data efficiency measures the average performance as a function of the data budget.
If $\text{AUC}_{k}$ is close to 1, this implies that performance saturates early with a small number of labeled examples; if closer to zero, this means little to no improvement is attainable from annotating additional examples.


\subsection{Predicting Data Efficiency}
\label{sec:predicting-data-efficiency}
Knowing the ground-truth data efficiency curve $f_k(n)$ and its $\text{AUC}_k$ for the task $k$ would inform the optimal fine-tuning data size. However, these measurements can only be made by fine-tuning the model at varying data budgets, necessitating access to a fully labeled dataset as well as sufficient computational resources.
An accurate estimate of task data efficiency could inform the shape of the data efficiency curve, which provide answers to valuable questions such as ``how many data points should I collect in order to achieve a desired level of performance?''
We therefore turn to devising a method for reliably estimating the data efficiency.
Notably, such a method is only valuable insofar as it does not require many labeled examples to perform estimation.

As far as we know, predicting the data efficiency of a task using cheap-to-compute metrics has not been explored before.
\cite{jiang2021characterizingstructuralregularitieslabeled} suggests that ``structural regularities'' of a set of data points, determined by how (a)typical the set is compared to the other examples, dictates the number of data points required to learn that class. Extending this idea, we hypothesize that a task with higher complexity due to heterogeneous or irregular data points has lower data efficiency.
We survey several past works studying the role of anomalous \citep{agarwal2022estimatingexampledifficultyusing,jiang2021characterizingstructuralregularitieslabeled,li2024rolelongtailknowledgeretrieval,pleiss2020identifyingmislabeleddatausing,siddiqui2022metadataarchaeologyunearthingdata}, long-tail \citep{feldman2020neuralnetworksmemorizewhy,hooker2021compresseddeepneuralnetworks}, or heterogeneous examples \citep{liu2024conflictaversegradientdescentmultitask,sener2019multitasklearningmultiobjectiveoptimization,shi2023reconreducingconflictinggradients,yu2020gradientsurgerymultitasklearning}, their occurrences surfaced by metrics involving model loss \citep{mindermann2022prioritizedtrainingpointslearnable}, predictions \citep{swayamdipta2020datasetcartographymappingdiagnosing}, or gradients \citep{agarwal2022estimatingexampledifficultyusing, paul2023deeplearningdatadiet}.
Drawing on this body of work, we repurpose the proposed metrics to characterize the target task's complexity (``task difficulty'') and the corresponding data efficiency.

\textbf{Baselines predictors of task difficulty}
As baseline metrics of task difficulty, we consider 1) the gradient $\text{L}_2$ norm and 2) the model's confidence.
The \textit{gradient norm} of the model's weights with respect to the model loss signals the magnitude of change in parameters required to shift the model's prediction to the target. Intuitively, learning a task with high gradient norm examples requires a larger change in the pre-trained model and therefore may require more data. The \textit{model confidence}, measured by averaging model probabilities assigned to its predictions, quantifies the degree of model certainty.
High model confidence may indicate familiarity with the task (\cite{shi2024detectingpretrainingdatalarge}), potentially indicating that learning the task to perfection only requires a few data points, or that only long-tail examples remain to be memorized, making learning data inefficient \citep{feldman2020neuralnetworksmemorizewhy, achille2020informationcomplexitylearningtasks,jiang2021characterizingstructuralregularitieslabeled}.



Unlike some of the past work tracking these metrics on the entire labeled dataset over the course of training \citep{agarwal2022estimatingexampledifficultyusing,paul2023deeplearningdatadiet,swayamdipta2020datasetcartographymappingdiagnosing}, we compute the per-sample metrics on a handful of data points at inference time and aggregate them to a single value for each task $k$, denoted $\texttt{grad\_norm}_k$ and $\texttt{conf\_avg}_k$ (see details in \cref{appendix:task-difficulty-metric-definition}).

\textbf{Our predictor: Gradient cosine similarity}
We ultimately find that these preexisting metrics do not serve as sufficiently reliable predictors of data efficiency in our experiments (\cref{sec:result}).
To address this shortcoming, we take inspiration from the multitask learning literature \citep{liu2024conflictaversegradientdescentmultitask,sener2019multitasklearningmultiobjectiveoptimization,shi2023reconreducingconflictinggradients,yu2020gradientsurgerymultitasklearning} that studies the \textit{conflicting gradient} problem, where the sample gradients from multiple tasks point in different directions, resulting in suboptimal multitask models.
To capture gradient conflict, it is typical to measure the cosine similarity between per-sample gradients of the model's weights with respect to the loss for different examples.
Unlike in the multitask learning methods that aim to minimize gradient conflict, we measure the degree of conflict \textit{within} a single task to estimate the task's learnability.
Specifically, we compute the median batch gradient cosine similarities of task examples (\cref{eq:cos-sim}):
\begin{equation}
\texttt{cos\_sim}_k = \text{median} \{ \cos(g_i, g_j)\ | (x_i, y_i), (x_j,y_j)\in B, i \neq j\}
\label{eq:cos-sim}
\end{equation}
where ($x_i$,$y_i$) and ($x_j,y_j$) are a pair of task data points in $B$, $g_i$, $g_j$ are the corresponding gradient of the weights with respect to the loss, and $\cos(g_i, g_j)$ measures the cosine similarity of two gradient vectors $\frac{g_i \cdot g_j}{||g_i|| ||g_j||}$.
In our experiments, we find that $\texttt{cos\_sim}_k$ computed on low-confidence task examples (\cref{eq:our-method}) is the most predictive of our data efficiency metric:
\begin{equation}
\text{\NameOfMeasurement} = \text{median} \{ \cos(g_i, g_j)\ | (x_i, y_i), (x_j,y_j) \in B, i \neq j, B \subseteq U_{0.1} \}
\label{eq:our-method}
\end{equation}
In practice, the batch of examples $B$ is sampled from $U_{0.1}$, the top 10\% of the low-confidence task examples. We compute \NameOfMeasurement with varying $U_{t}$, where $t \in \{0.1, 0.3, 0.5, 0.7, 1\}$, and find that the gradient conflict in lower confidence segment is more predictive of our notion of data efficiency (\cref{appendix:ablation-batch-and-confidence}).
Past work in active learning suggests that examples with high model uncertainty are the most informative for improving model performance (\cite{dredze-crammer-2008-active,hübotter2025efficientlylearningtesttimeactive}).
The role of low-confidence examples in model training may explain why the gradient alignment among such examples better predict how quickly a model improves with additional data. However, further theoretical analysis of this connection is left for future work.


\subsection{Mapping Task Difficulty to Data Efficiency}
\label{sec:task-difficulty-to-data-efficiency}

Recall that our overarching goal is to find a cheaply computable metric that correlates with our notion of data efficiency (\cref{sec:data-efficiency-definition}).
Given such a correlation, we might hope to be able to predict data efficiency and, consequently, estimate the corresponding data efficiency curve $\hat{f}_k(n)$ to propose the data budget required to reach a target performance.

To map one of the aforementioned task difficulty metrics to a task's data efficiency, we fit a simple linear regression model $\text{AUC}'=c * d + I$, where $\text{AUC}'$ is the predicted data efficiency, $d$ is one of $\texttt{grad\_norm}_k$, $\texttt{conf\_avg}_k$, $\texttt{cos\_sim}_k$, and \NameOfMeasurement, and $\{c,I\}$ are regression coefficients.
To test each of the metrics on the 30 downstream tasks introduced in \cref{sec:data-efficiency-variability}, we use a hold-one-out setting, in which all tasks except the held-out task $k$ are used in training to model the regressor. We then obtain the task $k$'s data efficiency prediction, $\text{AUC}_k'$, using the fitted regressor.

We can then use the estimated task data efficiency to produce a task-specific data efficiency curve $\hat{f}_k(n)$ such that its area under the curve (with both its axes normalized to 0 and 1) is precisely $\text{AUC}_k'$. However, there are multiple ways to model such a curve defined between 0 and 1 on both axes. Following the stated assumptions on the data efficiency curve (that it is a monotonically increasing curve defined between 0 and 1 on both axes, reaching the maximum performance within the maximum data budget of 5000), we consider in \cref{appendix:parametric-curve} various parametric functions $\hat{f}_k(n)$. After measuring how well each parametric family approximates the actual curve $f_k(n)$ in practice, we ultimately use the the following power function:
$$\hat{f}_k(n) = n^p, \text{ where } p = \frac{1-\text{AUC}_k'}{\text{AUC}_k'}$$
We include a high-level algorithm that maps a task difficulty metric to a fine-tuning data size for a target performance in \cref{appendix:cos-low-algorithm}.


\section{Experimental Setup}
\label{sec:experiment-setup}

\textbf{Metric calculation details.} To justify the use of task difficulty proxies from \cref{sec:measuring-data-efficiency} to predict task data efficiency, each metric should incur minimal annotation. Therefore, we require that $\texttt{grad\_norm}_k$, $\texttt{cos\_sim}_k$ and \NameOfMeasurement only use 32 annotated samples; deriving $\texttt{conf\_avg}_k$ is completely annotation-free. The per-sample $\texttt{grad\_norm}_k$ and $\texttt{conf\_avg}_k$ are aggregated to the task-level using median. Similarly, every pair-wise gradient cosine similarity for computing $\texttt{cos\_sim}_k$ and \NameOfMeasurement is aggregated using median. See \cref{tab:method-computation-overhead} for detailed computation overhead for each metric calculation. Throughout our experiments, $\text{AUC}_k$ are derived from the data efficiency curve with the data size in log-scale of base 2, as the performance shifts most notably in lower data budgets. 
Each of $\texttt{grad\_norm}_k$, $\texttt{conf\_avg}_k$, $\texttt{cos\_sim}_k$, and \NameOfMeasurement is used to fit the linear regressor to predict the task AUC for all 30 downstream tasks in a hold-one-out setting (\cref{sec:task-difficulty-to-data-efficiency}). We repeat the metric calculation on 32 examples end-to-end 7 times to measure the variance in AUC prediction.

As \NameOfMeasurement requires identifying low confidence examples, we run forward passes on at most 2500 \textit{unlabeled} examples to identify the top 10\% lowest confidence examples --- i.e., examples with the lowest \texttt{conf\_avg} --- and randomly sample 32 from the segment. In practice, this extra compute cost can be lower, as \NameOfMeasurement is robust to noise in the low confidence example selection (\cref{sec:discussion}). We run ablation studies calculating $\texttt{grad\_norm}_k$ and $\texttt{conf\_avg}_k$ on low confidence examples to fix the sample group as \NameOfMeasurement but do not find them to be very effective \cref{appendix:ablation-low-confidence-segment}.


\textbf{Using LoRA for gradient based metrics.} For calculating metrics requiring model gradients, $\texttt{grad\_norm}_k$, $\texttt{cos\_sim}_k$, and \NameOfMeasurement, we use rank-64 LoRA adaptors to store the gradients, which is both computation and memory efficient. LoRA gradients hold sufficient information needed to estimate task data efficiency compared to using full model gradients (\cref{appendix:ablation-gradient-projection}) and avoids the high cost for storing per-sample gradients in memory, which for Llama 3.1 8B Instruct model is $\approx \frac{32}{16} * \text{8} * 32\text{ examples} \approx 512 \text{GB}$, for a model loaded in half precision.

\textbf{Baselines.}
We use \texttt{base\_max} as an additional baseline that relies on the simple heuristic that there will be no performance increase before fine-tuning on the maximum budget (i.e., 5000 data points). This reflects an implicit assumption when the full annotation budget is used upfront before fine-tuning. As \texttt{base\_max} assumes static data efficiency curves across tasks, it serves as an upper bound of data efficiency prediction error.

\section{Experimental Results}
\label{sec:result}

\begin{figure}[b]
    \centering
    \includegraphics[width=\linewidth]{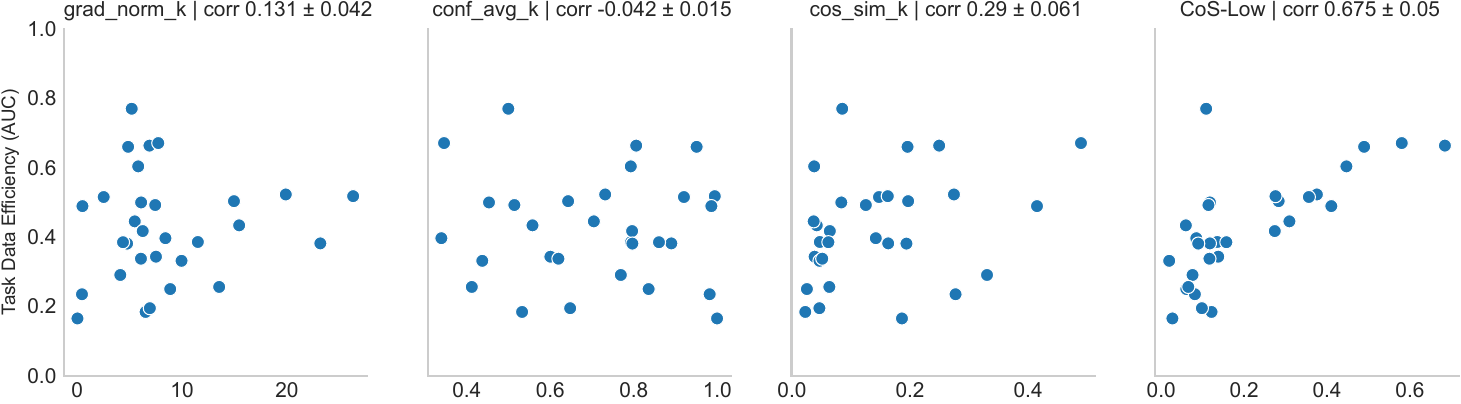}
    \caption{\NameOfMeasurement (right) shows the strongest relationship with task data efficiency among other task difficulty metrics. Each metric is compared with the ground-truth task data efficiency (y-axis) using Spearman's rank correlation.}
    \label{fig:data-efficiency-by-method}
\end{figure}

Across our experiments, \NameOfMeasurement shows the strongest performance in efficiently estimating data efficiency and reliably predicts the required fine-tuning data budget across the 30 downstream tasks. To evaluate performance, we report 1) the correlation between each metric and the task data efficiency, 2) absolute mean error of AUC prediction using each metric, and 3) analysis using \NameOfMeasurement to predict the fine-tuning data size prediction across desired performance levels.

\subsection{AUC Prediction Accuracy}

\NameOfMeasurement displays the strongest Spearman correlation (0.675) with task data efficiency among all metrics considered (\cref{fig:data-efficiency-by-method}). While past studies track model confidence or gradient magnitude throughout training to surface challenging examples or estimate model's generalization capability \citep{agarwal2022estimatingexampledifficultyusing,jiang2021characterizingstructuralregularitieslabeled,li2024rolelongtailknowledgeretrieval,pleiss2020identifyingmislabeleddatausing}, our results show that $\texttt{grad\_norm}_k$ and $\texttt{conf\_avg}_k$ computed at inference time do not display strong relationships with data efficiency. Consequently, predicting the task data efficiency using linear regression yields a statistically significant result only for \NameOfMeasurement (p-value $<$ 0.0002) and achieves the lowest prediction error (\cref{tab:overall-AUC-pred-accuracy-by-method}).
\begin{wraptable}{r}{7.4cm}
    \centering
        \begin{tabular}{c | c }
        \toprule
        \small
        Methods & Overall Abs. Mean Error \\
        \midrule
        \texttt{base\_max}              &    0.391 \\
        $\texttt{grad\_norm}_k$             &    0.130 $\pm$ 0.036 \\
        $\texttt{conf\_avg}_k$              &    0.133 $\pm$ 0.036 \\
        $\texttt{cos\_sim}_k$              &    0.124 $\pm$ 0.036 \\
        \NameOfMeasurement    &  \textbf{0.086 $\pm$ 0.030} \\
        \bottomrule
        \addlinespace
        \end{tabular}
    \caption{Mean absolute error in the AUC prediction using each method. \NameOfMeasurement (ours) has the lowest overall AUC prediction error (in \textbf{bold}) when compared with the ground-truth AUC of the 30 downstream tasks.}
    \label{tab:overall-AUC-pred-accuracy-by-method}
\end{wraptable}


While \NameOfMeasurement provides a reliable predictor for data efficiency, the relationship is much weaker for \texttt{cos\_sim\_k}, suggesting that gradient similarity provides a stronger signal among low confidence examples than from random examples. However, \texttt{grad\_norm\_k} or \texttt{conf\_avg\_k} computed on the same low confidence segment are not predictive of task data efficiency, which we further discuss in \cref{appendix:ablation-low-confidence-segment}.
Consequently, we can conclude that \NameOfMeasurement's performance stems from the fact that the cosine similarity of gradients is an especially useful signal when computed over low-confidence examples.

\subsection{Fine-tuning Data Size Prediction Accuracy}

To translate the observed performance of \NameOfMeasurement into tangible cost savings, we run a cost analysis comparing our task-specific data efficiency prediction method and alternative task-agnostic approaches for finding the optimal fine-tuning data size for the desired performance level. In practice, one can incrementally annotate and repeatedly fine-tune the model until the target performance is reached (``incremental annotation''), or annotate the full dataset (up to 5000 examples) and run a single fine-tuning (``maximum annotation''). \NameOfMeasurement serves as an in-between approach, where we first fine-tune with the predicted data size, then only train further with incremental annotation approach if the desired performance has not been reached.

We model the fine-tuning cost $C$ as a fixed amount per training run, as the cost of repeated training include access to training resources and human oversight, which does not scale linearly with the dataset size. $A$ denotes the per-example cost of annotation. We assume the number of fine-tuning examples required to reach near-human-level performance is one of 50, 100, 200, 500, 1000, 2500, and 5000. The ground truth data size required to reach the desired performance levels are empirically measured from the 30 downstream tasks. 


As shown in \cref{tab:excess-annot-and-compute-by-method}, \NameOfMeasurement's approach balance the trade-off between the ``maximum annotation'' and ``incremental annotation'' approaches, achieving relatively low excess annotation and few extra training runs compared to either extreme. With the cost of fine-tuning as a function of annotation and compute, practitioners can assess the given annotation and compute cost ratio to adopt a more desirable option. We include further analysis on our method's prediction error in \cref{appendix:fine-tuning-data-size-pred-error}. 

\begin{table}[H]
    \centering
    \resizebox{1.\textwidth}{!}{%
        \begin{tabular}{c | c c | c c | c c}
        \toprule
        \multirow{2}{*}{Desired Perf.} & \multicolumn{2}{c}{Incremental Annotation} & \multicolumn{2}{c}{Maximum Annotation} & \multicolumn{2}{c}{Ours} \\
         & Extra Annot. &  Extra Training & Extra Annot. &  Extra Training & Extra Annot. &  Extra Training \\
        \midrule
        70\% &  0 & 3$C$  & 3860$A$ & 0 &  219$A$ & 1$C$ \\
        80\% &  0 & 4$C$  & 3209$A$   & 0 &  748$A$	& 1$C$ \\
        90\% &  0 & 5$C$ & 2602$A$  & 0  & 701$A$	& 1$C$ \\
        95\% &  0 & 5$C$ & 1699$A$  & 0  & 1115$A$   & 1$C$ \\
        \bottomrule
        \addlinespace
        \end{tabular}
    }%
    \caption{Incremental annotation leads to 5 additional fine-tuning runs on average to reach 95\% of the human-level performance. Maximum annotation wastes annotations across all desired performance levels. Even when \NameOfMeasurement approach underestimates the data size required, necessitating incremental annotation, it only requires one extra fine-tuning run on average and much lower wasted annotation cost.}
    \label{tab:excess-annot-and-compute-by-method}
\end{table}

\section{Ablation Studies and Further Analysis}
\label{sec:discussion}

\begin{table}
    \centering
    \resizebox{0.9\textwidth}{!}{%
    \small
        \begin{tabular}{c | c c c}
        \toprule
        \small
        Coefficient & Llama 3.1 8B-Instruct & Mistral 7B-Instruct v.03 &  Qwen 2.5 14B-Instruct\\
        \midrule
        \NameOfMeasurement & 0.545$\pm$0.005     &    0.797$\pm$0.012  & 0.526$\pm$0.025\\
        Intercept          & 0.310$\pm$0.002     &    0.357$\pm$0.002  & 0.305$\pm$0.003\\
        \bottomrule
        \addlinespace
        \end{tabular}
    }%
    \caption{Regression coefficient to map \NameOfMeasurement to data efficiency varies across model families.}
    \label{tab:regression-coefficient-across-models}
\end{table}

\paragraph{Does \NameOfMeasurement display strong correlation with data efficiency across models families of varying sizes?}

To validate the robustness of our results across model families and sizes, we replicate our experiments on Mistral 7B Instruct v0.3 \citep{jiang2023mistral7b} and Qwen 2.5 14B Instruct \citep{bai2023qwentechnicalreport}. We find that \NameOfMeasurement remains the strongest metric for data efficiency prediction across model families (\cref{fig:model-families-correlation}).
However, the regression coefficients used to map \NameOfMeasurement to task data efficiency for one model family cannot be reused for another, as these vary across model families (\cref{tab:regression-coefficient-across-models}).
While this poses a potential challenge, we note that the regression weights only need to be computed once for each model and can be reused indefinitely for new downstream tasks. Alternatively, the weights can be shared collaboratively within a community to support efficient training.
Training setup and \NameOfMeasurement correlation with data efficiency for these model families are discussed in \cref{appendix:method-to-model-family}.

\paragraph{Can \NameOfMeasurement reliably predict data efficiencies of out-of-distribution tasks?}

To test the generalizability of \NameOfMeasurement beyond the original 30 downstream tasks (primarily classification and multiple-choice QA), we extend our experiments to 10 out-of-distribution (OOD) tasks not included in the original set. The OOD tasks comprise two multi-task dataset collections, four generation tasks, and four domain-specific downstream tasks (see \cref{appendix:method-to-task-categories} for details). We first validate that \NameOfMeasurement continues to display meaningful correlation with task data efficiency among the OOD tasks. We then test the reusability of the regression coefficients learned from the original 30 downstream tasks (``held-in'' tasks) to predict the OOD tasks' data efficiencies.

\begin{figure}[t!]
    \begin{minipage}[c]{0.50\textwidth}
    \centering
    \small
    \vspace{0pt}%
    \begin{tabular}{c c c}
        \toprule
        \small
        Task & Corr.\ with AUC & Abs.\ mean err \\
        \midrule
        OOD              & 0.727 $\pm$ 0.02 & 0.109 $\pm$ 0.05 \\
        OOD + Held-in       & 0.712 $\pm$ 0.04 & 0.086 $\pm$ 0.03 \\
        Held-in    & 0.675 $\pm$ 0.05 & 0.086 $\pm$ 0.03 \\
        \bottomrule
    \end{tabular}
    \captionof{table}{\NameOfMeasurement's correlation with task data efficiency and the AUC prediction error for the held-in and OOD tasks.}
    \label{fig:ood-cos-low-stats}
    \vspace{1em}
    \centering
    \small
    \begin{tabular}{lcc}
        \toprule
        \small
        \NameOfMeasurement  & Average & Median \\
        \midrule
        $d=0$ & 0.210 $\pm$ 0.066 & 0.126 \\
        $d=50$ & 0.108 $\pm$ 0.046& 0.063 \\
        $d=500$ & 0.095 $\pm$ 0.057 & 0.035 \\
        \bottomrule
    \end{tabular}
    \captionof{table}{\NameOfMeasurement computed on model fine-tuned with $d$-many target data points across 30 tasks. The decrease in \NameOfMeasurement indicate reduced task data efficiency with additional fine-tuning.}
    \label{tab:cos-low-across-budget}
    \end{minipage}%
    \hspace{0.04\textwidth}%
    \begin{minipage}[c]{0.48\textwidth}
        \centering
        \vspace{0pt}%
        \includegraphics[width=\linewidth]{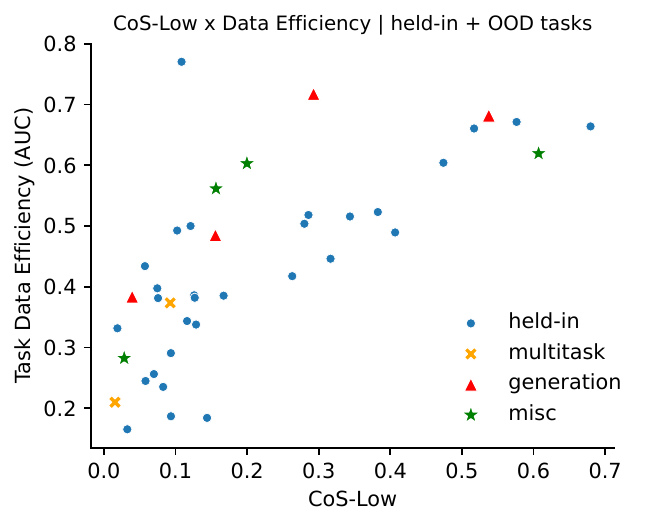}
        \caption{\NameOfMeasurement and task AUC across the 30 held-in (the original downstream tasks) and 10 OOD tasks.}
        \label{fig:30-vs-ood}
    \end{minipage}
\end{figure}

Our finding shows that \NameOfMeasurement is indeed a reliable proxy for task data efficiency across a more complex set of tasks, including multi-task and generation tasks (Spearman's rank correlation of 0.759, \cref{fig:ood-cos-low-stats}). Moreover, once learned, the regression coefficients are reusable to predict data efficiency of unseen downstream tasks, as OOD tasks generally exhibit a similar linear relationship between \NameOfMeasurement and data efficiency (\cref{fig:30-vs-ood}). However, the mismatch between accuracy-based data efficiency curves in the 30 held-in tasks and F1-based curves in generation tasks may introduce errors in the prediction (e.g. the outlier among generation tasks in \cref{fig:30-vs-ood}), requiring further study.

\paragraph{How does \NameOfMeasurement evolve as the model is further trained on the target task?}

We next investigate whether \NameOfMeasurement can track changes in task data efficiency as the base model is trained on the target task. Empirically, incremental data points add less value as the model approaches saturation (\cref{appendix:ablation-data-budget-10k}, \cref{tab:raw-acc-gain}), implying the task data efficiency drops as the model is fine-tuned on the target task. To investigate the evolution of \NameOfMeasurement over the course of training, we compute \NameOfMeasurement on the checkpoints trained with data budget $d = \{0, 50, 500\}$, with $d=0$ being the value reported in \cref{sec:result}, computed on the base model.


We find that \NameOfMeasurement indeed tends to decrease as the base model is fine-tuned with more data (\cref{tab:cos-low-across-budget}). This suggests that \NameOfMeasurement evolves dynamically to reflect the model's degree of saturation. Applying this insight, \NameOfMeasurement has a potential use case in guiding black-box model checkpoint selection from the same model family. One can prioritize checkpoints with the highest or the lowest \NameOfMeasurement, as the extreme values indicate either rapid model improvement with low-budget fine-tuning or task saturation, possibly signaling higher performance on the target task.


\paragraph{When is \NameOfMeasurement's core assumption not met?}

While we empirically confirmed our our assumption that performance reaches a known human-level performance within the given maximum budget, it might may not always be true. For instance, MMLU (multitask accuracy across 57 subjects, spanning various topics from algebraic math to philosophy) \citep{hendrycks2021measuringmassivemultitasklanguage} and MedMCQA (more complex dataset, containing medical entrance exam covering 21 medical subjects and 2,400 healthcare topics) \citep{pal2022medmcqalargescalemultisubject} are tasks for which performance remains around $<75\%$ of human-level with 10,000 training examples. 

For such tasks (where the data efficiency curves of these tasks do not follow the proposed $n^p$ curve) we observe that the error in fine-tuning data size prediction grows larger. This failure mode of \NameOfMeasurement highlights the difficulty of estimating the point of performance saturation for a given task, which may be below the human-level performance. Using human-level performance as a proxy for maximum attainable performance may overestimate the true saturation point of the model, adding noise to the prediction.


\paragraph{Is the maximum budget of 5000 data points sufficient?}

As we use the higher of human-level performance and the maximum observed accuracy within 5,000 examples to approximate the model's maximum attainable performance, it is important to confirm that the performance within this data budget is sufficiently close to the true maximum. To establish this cut-off, we first verify that performance gains beyond 5,000 data budget are marginal across the 30 tasks (\cref{appendix:ablation-data-budget-10k}, \cref{tab:raw-acc-gain}). In addition, we show the impact of increasing the maximum data budget to 10,000 (by constraining the set of datasets to those that support this data budget) and rerunning the experiments does not add significant changes to our findings (\cref{appendix:ablation-data-budget-10k}, \cref{tab:10k-data-budget}). This result highlights that \NameOfMeasurement correlates most strongly with earlier performance improvements at smaller budgets, rather than the small gains observed at the tail end.

\paragraph{How robust is \NameOfMeasurement to Sample Size and the Low-Confidence Segment?}

Throughout our experiments, we select 32 task data samples among the top 10\% of low-confidence examples to calculate \NameOfMeasurement. While we have demonstrated its high correlation with our data efficiency measure, we explore how sensitive our method is to the choice of sample size and low-confidence segment. We vary the sample size and the low-confidence segment and examine 1) the correlation between the newly computed \NameOfMeasurement and task data efficiency, and 2) the mean absolute AUC prediction error.

We randomly select 4, 8, and 16 examples among the low-confidence segment of the downstream task to compute \NameOfMeasurement and use them to predict task data efficiency. 
We find that the relationship between \NameOfMeasurement and the task AUC becomes weaker (\cref{fig:cos-sim-batch-size} in \cref{appendix:ablation-batch-and-confidence}) and the overall AUC prediction error increases with smaller batch size (\cref{tab:error-by-batch-size} in \cref{appendix:ablation-batch-and-confidence}). However, the AUC prediction still has a statistical significance (p-value $<$0.05) using sample size of 8 or 16, suggesting our method is reasonably robust to the choice of sample size.

Another key step in computing \NameOfMeasurement is the selection of datapoints in the ``low confidence segment'' of the task dataset. To measure the sensitivity to datapoint selection from the low-confidence segment, we sample task data points from the top 30\%, 50\%, and 70\% of the low-confidence segment. Notably, sampling examples from the top 30\% or even 50\% of low-confidence segment still produces a Spearman's rank correlation greater than 0.5 with the task data efficiency (\cref{fig:cos-sim-conf-seg} in \cref{appendix:ablation-batch-and-confidence}) and results in statistically significant AUC prediction (\cref{tab:error-by-conf} in \cref{appendix:ablation-batch-and-confidence}). This result indicates that our method can perform well without needing to scan the entire dataset to identify the lowest-confidence examples.

\section{Related Work}

\textbf{Data efficiency} In the context of pre-trained LLMs, past work (\cite{aghajanyan2020intrinsicdimensionalityexplainseffectiveness,brown2020languagemodelsfewshotlearners,sanh2022multitaskpromptedtrainingenables,wei2022finetunedlanguagemodelszeroshot,zhou2023limaalignment}) demonstrates that knowledge is mostly learned during the pretraining phase, allowing for effective knowledge transfer during fine-tuning. However, learning long-tail knowledge requires memorization and typically requires more data (\cite{achille2020informationcomplexitylearningtasks,feldman2020neuralnetworksmemorizewhy,hooker2021compresseddeepneuralnetworks, jiang2021characterizingstructuralregularitieslabeled}).
\cite{zhang2024scalingmeetsllmfinetuning} quantifies the impact of fine-tuning data size on the downstream performance to establish a fine-tuning scaling law.
For various data efficiency predictors discussed, we take inspiration from multi-task learning and active learning literature. Multi-task learning research \cite{yu2020gradientsurgerymultitasklearning, liu2024conflictaversegradientdescentmultitask,sener2019multitasklearningmultiobjectiveoptimization,shi2023reconreducingconflictinggradients,yu2020gradientsurgerymultitasklearning} has introduced the concept of \textit{conflicting gradients} among more than two tasks, causing convergence difficulties. Active learning approaches aim to choose which unlabeled training samples should be selected for labeling, using statistics such as model uncertainty (\cite{dredze-crammer-2008-active,hübotter2025efficientlylearningtesttimeactive}).



\textbf{Task difficulty}
Past work that aims to measure task difficulty often examines sample-level statistics tracked over training. Some work tracks the variance of the model confidence (\cite{swayamdipta2020datasetcartographymappingdiagnosing}) or per-sample gradients (\cite{agarwal2022estimatingexampledifficultyusing}) during training to surface hard or ambiguous examples. \cite{pleiss2020identifyingmislabeleddatausing,siddiqui2022metadataarchaeologyunearthingdata} study data taxonomy (e.g. typical, atypical, challenging, mislabeled, etc.) by observing a data point's learning curve during training. Other work aims to select a subset of more challenging or useful examples to learn the task more data-efficiently (\cite{mindermann2022prioritizedtrainingpointslearnable,paul2023deeplearningdatadiet}). These works observe that difficult examples tend to be highly ambiguous or without consistent labels, impacting the rate of learning. We refer to these works and use sample-level difficulty proxies to compute task-level difficulty, but our setting differs because we cannot measure training trajectories without performing fine-tuning.

\section{Conclusion}
In our work, we introduce a notion of task data efficiency using the AUC of data efficiency curve as the fine-tuning data size increases. We empirically show that data efficiency can vary dramatically across downstream tasks and  aim to predict data efficiency by exploring several measures of task difficulty. Our chosen method leverages the median gradient cosine similarity of low-confidence examples, \NameOfMeasurement, and can efficiently estimate the task data efficiency using as few as 32 task examples. Finally, we show that using our method to find the optimal data size for a desired performance level can save unnecessary annotation or fine-tuning cost incurred when using simple heuristics.

One future direction of our work is to extend our method to generation tasks using non-accuracy based metrics (e.g., BLEU~\citep{papineni-etal-2002-bleu} or ROUGE~\citep{lin-2004-rouge} scores, or even LLM-as-a-judge evaluation~\citep{gu2025surveyllmasajudge}). Another direction is establishing a more rigorous relationship between model evolution after fine-tuning and low-confidence training samples’ gradient cosine similarity. Currently, our work focuses on practical implementation with high-level theoretical justification. Lastly, we assume either the highest observed performance within the data budget or human-level performance is the maximum attainable performance on any given model. In future research, metrics derived from model internals, including the ones considered in our work for task difficulty estimation, can be used to check the degree of model saturation and find the model-specific upper-bound.

\bibliographystyle{iclr2026_conference}
\bibliography{references}

\newpage
\appendix
\section{Downstream Task Overview}
\label{appendix:downstream-task-details}


We select 30 downstream tasks that span multiple domains including healthcare, law, finance, safety, and other domains requiring natural language reasoning ability. All but three tasks have at least 2500 training examples (\textit{Temporal\_sequences} \citep{srivastava2023imitationgamequantifyingextrapolating} has 800, \textit{RTE} \citep{wang2020supergluestickierbenchmarkgeneralpurpose} 2241, \textit{Overruling} \citep{zheng2021doespretraininghelpassessing} 1920). Since our data efficiency metric---task AUC---requires evaluating model performance with up to 5000 fine-tuning examples, we extrapolate the performance for these tasks by assuming their peak performance at the maximum available data size is comparable to the performance at fine-tuning data size of 5000.

\begin{figure}[hb]
    \centering
    \includegraphics[width=0.6\linewidth]{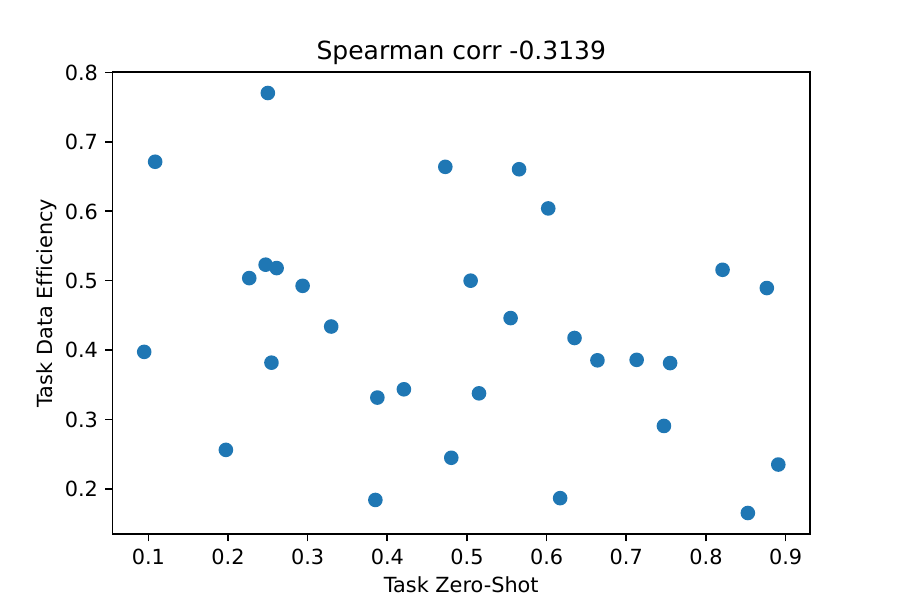}
    \caption{Relationship between zero-shot accuracy and task data efficiency. While higher zero-shot accuracy of tasks close to performance saturation may indicate lower task data accuracy, the relationship is not consistent (Spearman rank correlation of -0.3139).}
    \label{fig:zero-shot-perf}
\end{figure}

\cref{tab:downstream-task-acc-auc} provides a high-level overview of each task, including its zero-shot accuracy, maximum performance after fine-tuning, maximum attainable performance (defined as the greater of known human-level accuracy or the best fine-tuned performance with the 5000-example data budget), and the task data efficiency. We show that neither high or low task zero-shot performance consistently predicts task data efficiency in \cref{fig:zero-shot-perf}, highlighting that estimating downstream task data efficiency is a non-trivial problem. Below, we categorize the tasks by their relevant domains and briefly describe each.

\textbf{Medical}

\textit{Ade\_corpus\_v2\_classification} \citep{GURULINGAPPA2012885} consists of medical statements indicating the presence of an adverse drug event (ADE=1 or 0), designed to support the extraction of drug-related adverse effects from medical case reports. \textit{MedMCQA} \citep{pal2022medmcqalargescalemultisubject} is a multiple-choice question dataset derived from a real-world medical entrance exam covering 21 medical subjects and 2,400 healthcare topics.

\textbf{Law}

Overruling \citep{zheng2021doespretraininghelpassessing} comprises extracted sentences from legal opinions, a subset of which overrule a prior decision (label=1, 0 otherwise).

\textbf{Intent Detection}

Banking77 \citep{casanueva-etal-2020-efficient} consists of online banking queries labeled with one of 77 predefined user intent categories, supporting intent classification in the financial service domain.
\textit{ToxicChat} \citep{lin2023toxicchatunveilinghiddenchallenges} consists of user prompts collected from the Vicuna online demo, annotated for toxicity in the user prompts.
\textit{Circa} \citep{louis-etal-2020-id} presents brief question-answer dialogues with ambiguous responses and crowd-sourced ground-truth labels indicating the underlying intention of the ambiguous answer.

\textbf{World Knowledge}

\textit{CommonsenseQA} \citep{talmor-etal-2019-commonsenseqa} evaluates commonsense reasoning ability requiring prior knowledge across a range of target concepts.
\textit{MMLU} \citep{hendrycks2021measuringmassivemultitasklanguage} assesses multitask accuracy across 57 subjects, spanning various topics from algebraic math to philosophy.
\textit{Sports\_understanding} \citep{srivastava2023imitationgamequantifyingextrapolating} examines general understanding of sports by presenting plausible or implausible statements related to sports, given specific actions in sports and names of athletes.
\textit{Hyperbaton} \citep{srivastava2023imitationgamequantifyingextrapolating} tests the ability to identify the correct order of adjectives in given text.

\textbf{Logical Deduction and Reasoning}

\textit{Boolean\_expressions} and \textit{Web\_of\_lies} \citep{srivastava2023imitationgamequantifyingextrapolating} consist of nested boolean logic, presented either in formal notation or natural language, that evaluate to True or False.
\textit{Formal\_fallacies\_syllogisms\_negation} \citep{srivastava2023imitationgamequantifyingextrapolating} assesses the ability to distinguish between deductively valid and invalid arguments given a premise and corresponding argument.
\textit{Object\_counting} \citep{srivastava2023imitationgamequantifyingextrapolating} evaluates the ability to count simple objects described in a sentence while ignoring irrelevant distractors.
\textit{Temporal\_sequences} \citep{srivastava2023imitationgamequantifyingextrapolating} requires deduction over a sequence of temporally ordered events.
\textit{Tracking\_shuffled\_objects} \citep{srivastava2023imitationgamequantifyingextrapolating} tests the ability to track object ownership as the object is transferred among multiple individuals in a sequence of actions.

\textbf{Classic Natural Language Inference}

\textit{ANLI} \citep{nie-etal-2020-adversarial} and \textit{MNLI} \citep{wang2019gluemultitaskbenchmarkanalysis}, and \textit{RTE} \citep{wang2020supergluestickierbenchmarkgeneralpurpose} are natural language inference (NLI) benchmarks, each consisting of a premise and a hypothesis, with their relationship categorized as entailment, contradiction, or neutral. \textit{ANLI} is constructed via adversarial human-and-model-in-the-loop procedure; \textit{MNLI} consists of text extracted from speech, fiction, government speech; and RTE comprises news and Wikipedia texts.

\textbf{Miscellaneous Natural Language Understanding}

\textit{QQP} \citep{wang2019gluemultitaskbenchmarkanalysis} and \textit{MRPC} \citep{wang2019gluemultitaskbenchmarkanalysis} assess semantic equivalence between pairs of sentences extracted from the Quora discussion forum and online news respectively.
\textit{SST-2} \citep{wang2019gluemultitaskbenchmarkanalysis} is a sentiment classification task based on movie reviews.
\textit{Fig-QA} \citep{liu2022testingabilitylanguagemodels} evaluates the ability to interpret figurative language given human-written creative metaphors.
\textit{WiC} \citep{wang2020supergluestickierbenchmarkgeneralpurpose} is a word sense disambiguation task determining if a polysemous word has the same meaning in two different text snippets.

\textbf{Reading Comprehension}

\textit{QuAIL} \citep{Rogers_Kovaleva_Downey_Rumshisky_2020} and \textit{RACE} \citep{lai2017racelargescalereadingcomprehension} are multiple-choice reading comprehension tasks. \textit{QuAIL} consists of texts extracted from fiction, news articles, blogs, and the Quora forum. \textit{RACE} is based on English exam passages designed for Chinese students aged between 12 and 18; in our experiments, we use the subset containing high-school level passages. \textit{BoolQ} \citep{wang2020supergluestickierbenchmarkgeneralpurpose} consists of a short passage paired with a yes-or-no question related to the passage. \textit{QNLI} \citep{wang2019gluemultitaskbenchmarkanalysis} assesses whether the answer to a question can be inferred from a given paragraph extracted from Wikipedia.

\textbf{Visual and Spatial Reasoning}

\textit{MNIST\_ascii} \citep{srivastava2023imitationgamequantifyingextrapolating} is a multi-label classification task based on the original MNIST dataset, where digits from 0 to 9 are rendered in ASCII string format rather than images.
\textit{Reasoning\_about\_colored\_objects} \citep{srivastava2023imitationgamequantifyingextrapolating} assesses the ability to understand spatial relationships by interpreting visual descriptions of scenes involving colored objects.

\begin{table}[ht]
    \centering
    \resizebox{1.\textwidth}{!}{%
        \begin{tabular}{c | c c c c}
        \toprule
        \multirow{2}{*}{Task} & \multicolumn{2}{c}{Model Accuracy} & \multirow{2}{*}{Max Attain. Acc.} & \multirow{2}{*}{Task AUC}\\
          & Zero-shot & Max Fine-tuned & & \\
        \midrule
        BoolQ & 0.85 & 0.90 & 0.90 & 0.165 \\
        ANLI & 0.39 & 0.74 & 0.92 & 0.184 \\
        WiC & 0.62 & 0.86 & 0.86 & 0.186 \\
        SST-2 & 0.89 & 0.95 & 0.98 & 0.235 \\
        Formal\_fallacies\_syllogisms\_negation & 0.48 & 0.99 & 0.99 & 0.245 \\
        Tracking\_shuffled\_objects & 0.20 & 0.95 & 1.00 & 0.256 \\
        MRPC & 0.75 & 0.88 & 0.88 & 0.291  \\
        Reasoning\_about\_colored\_objects & 0.39 & 0.94 & 1.00 & 0.332 \\
        Web\_of\_lies & 0.52 & 1.00 & 1.00 & 0.338 \\
        MedMCQA & 0.42 & 0.79 & 0.90 & 0.343 \\
        QQP & 0.76 & 0.86 & 0.86 & 0.381  \\
        MMLU & 0.25 & 0.65 & 0.90 & 0.382  \\
        Sports\_understanding & 0.66 & 0.99 & 1.00 & 0.386 \\
        Boolean\_expressions & 0.71 & 0.99 & 1.00 & 0.397 \\
        MNIST\_ascii & 0.09 & 0.94 & 0.98 & 0.397 \\
        MNLI & 0.64 & 0.87 & 0.92 & 0.417 \\
        Banking77 &  0.33 & 0.93 & 0.93 & 0.434\\
        Fig\_qa & 0.55 & 0.95 & 0.95 & 0.446 \\
        Toxicchat0124 & 0.88 & 0.97 & 1.00 & 0.489 \\
        QuAIL & 0.29 & 0.84 & 0.84 & 0.492 \\
        RACE & 0.50 & 0.84 & 0.85 & 0.500 \\
        CommonsenseQA & 0.23 & 0.80 & 0.89 & 0.504  \\
        Overruling & 0.82 & 0.97 & 0.97 & 0.516  \\
        RTE & 0.26 & 0.88 & 0.94 & 0.518 \\
        Object\_counting & 0.25 & 0.97 & 0.97 & 0.523 \\
        Hyperbaton & 0.60 & 1.00 & 1.00 & 0.604 \\
        QNLI & 0.57 & 0.93 & 0.93 & 0.660 \\
        Ade\_corpus\_v2\_classification & 0.47 & 0.95 & 0.95 & 0.664  \\
        Circa & 0.11 & 0.91 & 0.92 & 0.671  \\
        Temporal\_sequences & 0.25 & 1.00 & 1.00 & 0.770 \\
        \bottomrule
        \addlinespace
        \end{tabular}
    }%
    \caption{Downstream task's zero-shot accuracy, maximum accuracy after fine-tuning, maximum attainable accuracy (greater of the the human-level performance or the maximum fine-tuned accuracy), and task data efficiency metric (AUC). The tasks are sorted in the order of ascending task AUC, just as in \cref{fig:overall-task-data-efficiency}}
    \label{tab:downstream-task-acc-auc}
\end{table}

\section{Task Difficulty Metric Definitions}
\label{appendix:task-difficulty-metric-definition}

We compute $\texttt{grad\_norm}_k$ as per-sample $\text{L}_2$ gradient norm of weights with respect to the cross-entropy loss (\cref{eq:grad-norm}), aggregated to the task-level:

\begin{equation}
\texttt{grad\_norm}(x_i,y_i) = || \nabla_w L(x_i, y_i) ||
\label{eq:grad-norm}
\end{equation}
\begin{equation}
\texttt{grad\_norm}_k = \text{median}\{\texttt{grad\_norm}(x_i, y_i) \text{ }|\text{ } (x_i,y_i)\in B\} 
\label{eq:grad-norm-task}
\end{equation}
where ($x_i$, $y_i$) is an $i$-th input and corresponding target label with length $T$, from a randomly sampled set of task data points $B$. $L$ is the cross-entropy loss $ -\frac{1}{T} \sum_{t=0}^T\log P[y_{it}]$, and $P[y_{it}]$ is the probability assigned by the model to the $t$-th target label.

$\texttt{conf\_avg}_k$ is computed by averaging the model probabilities assigned to the predicted target $y'_i$ generated using greedy decoding (\cref{eq:conf-avg}). We then aggregate them to the task-level using median (\cref{eq:conf-avg-task}).
\begin{equation}
\texttt{conf\_avg}(x_i,y_i) = \frac{1}{T} \sum_{t=1}^T P[y'_{it}]
\label{eq:conf-avg}
\end{equation}
\begin{equation}
\texttt{conf\_avg}_k = \text{median}\{\texttt{conf\_avg}(x_i,y_i) \text{ }|\text{ } (x_i,y_i)\in B\} 
\label{eq:conf-avg-task}
\end{equation}

Note that $T$ is the length of the target label $y$ and is known in advance because our setup mainly considers short generation tasks.

\section{Fine-tuning Setup}
\label{appendix:fine-tuning-detail}


To measure task data efficiencies, we run full model fine-tuning on Llama 3.1 8B Instruct, Mistral 7B Instruct v0.3 and Qwen 2.5 14B Instruct on each of the 30 downstream tasks (results in \cref{sec:data-efficiency-variability}). All experiments are conducted using two Nvidia H100 GPUs for the 8B and 7B models, four for the 14B model, on a high-performance compute cluster. We use a warmup ratio of 0.1, an effective batch size of 32, a learning rate of 1e-5, and a cosine learning rate scheduler. Models are trained for a maximum of 500 steps, and the reported fine-tuned performance corresponds to the checkpoint with the lowest evaluation loss within the 500 steps. We use early stopping with a patience of 20, terminating training if the evaluation loss does not improve over 20 consecutive logging steps. Training examples with sequence lengths exceeding 2048 tokens are filtered out. All training runs use a fixed random seed for reproducibility.

Rounds of fine-tuning and evaluation with varying fine-tuning data sizes (50, 100, 200, 500, 1000, 2500, 5000) use the same test split within the same task. The test set contains up to 5000 examples. Validation set sizes are capped at 20\% of the corresponding training size (e.g., a training set of 50 examples use a validation set of at most 10 examples) to reflect realistic low-resource fine-tuning conditions.

\section{Parametric Curve to Model Data Efficiency}
\label{appendix:parametric-curve}

\begin{figure}[h]
    \centering
    \begin{subfigure}{1\textwidth}
        \centering
        \includegraphics[width=0.8\linewidth]{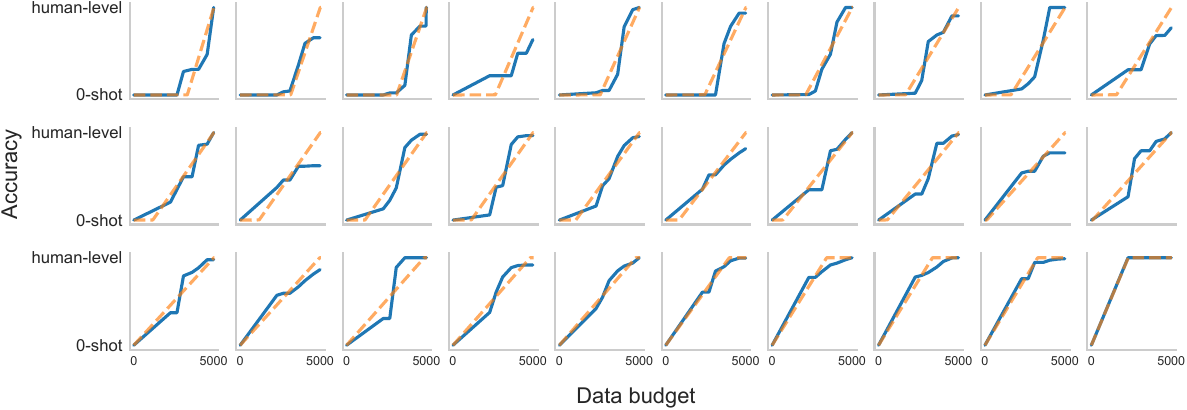}
        \caption{Actual data efficiency curve and the predicted data efficiency curve using linear function.}
        \label{fig:actual-vs-pred-linear}
        \vspace{2ex}
        \includegraphics[width=0.8\textwidth]{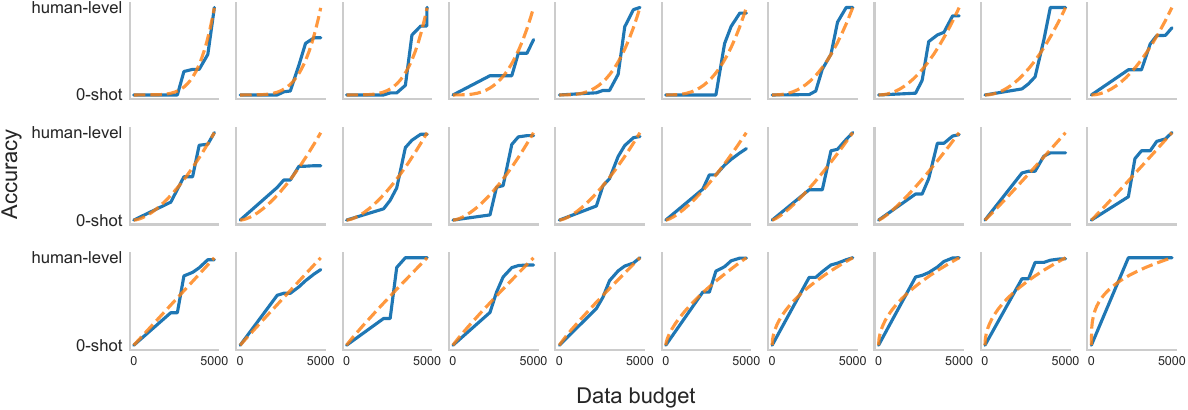}
        \caption{Actual data efficiency curve and the predicted data efficiency curve using power function.}
        \label{fig:actual-vs-pred-power}
    \end{subfigure}
    \caption{The actual data efficiency curve compared with both power and linear functions.}
    \label{fig:actual-vs-pred-curve-fit}
\end{figure}

As discussed in \cref{sec:task-difficulty-to-data-efficiency}, we map the predicted task data efficiency $\text{AUC}_k'$ to a task-specific data efficiency curve $\hat{f}_k(x)$ using a power function $x^p$, where $p = \frac{1-\text{AUC}_k'}{\text{AUC}_k'}$, to model the relationship between fine-tuned performance and fine-tuning data size. In this section, we show an alternative approach using a piecewise linear function \cref{eq:piecewise-linear} to map $\text{AUC}_k'$ to the data efficiency curve $\hat{f}_k(x)$:

\begin{equation}
\hat{f}_k(x) =
\begin{cases}
\min \{ \frac{1}{2(1-\text{AUC}_k')} * x , 1\} & \text{AUC}_k' \geq 0.5\\
\max \{\frac{1}{2\text{AUC}_k'} (x - 1) + 1, 0\} & \text{AUC}_k' < 0.5\\
\end{cases}
\label{eq:piecewise-linear}
\end{equation}

where $x$ is the percentage of the data budget (i.e., data size normalized between 0 and 1). We compare the fit of the predicted data efficiency curves $\hat{f}_k(x)$, estimated using either the power function or the piecewise linear function, with the original data efficiency curves $f_k(x)$ (\cref{fig:actual-vs-pred-curve-fit}).

The absolute error of the fit for the power function is slightly higher, at 8.47\% error, whereas the piecewise linear function has 8\% error. Despite the marginal difference, we choose the power function in our analyses as it better captures the gradual performance improvements in the low-data regime, whereas the piecewise linear function introduce a sharp transition.

\section{Fine-tuning Data Size Prediction Error}
\label{appendix:fine-tuning-data-size-pred-error}

\begin{figure}[ht]
    \centering
    \includegraphics[width=0.5\textwidth]{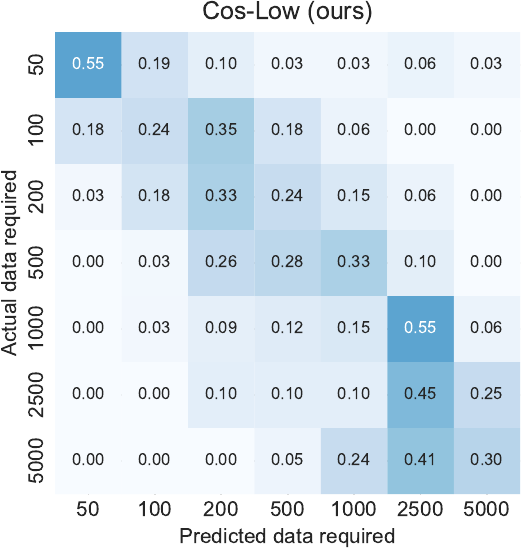}
    \caption{Actual vs. Predicted fine-tuning data size across all desired performance levels between 40\% and 95\%.} 
    \label{fig:pred-vs-actual-cm}
\end{figure}

\cref{fig:pred-vs-actual-cm} illustrates \NameOfMeasurement's data size prediction error across varying desired performance levels of 40\%, 50\%, 60\%, 70\%, 80\%, 90\%, and 95\%. \cref{fig:pred-vs-actual-cm} illustrates that our method is able to identify cases where only a small number of fine-tuning examples are sufficient (illustrated by the darker blue diagonal squares in the upper-left corner of \cref{fig:pred-vs-actual-cm}).

\section{Ablation Studies}

\subsection{Sample Size and Confidence Segment}
\label{appendix:ablation-batch-and-confidence}

\begin{figure}[h]
    \centering
    \begin{subfigure}{1\textwidth}
        \centering
        \includegraphics[width=0.8\linewidth]{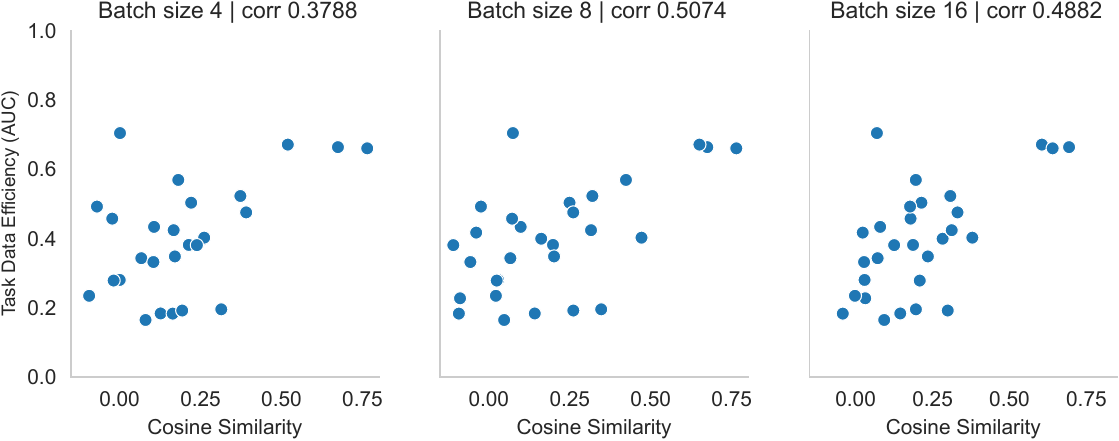}
        \caption{Task data efficiency (AUC) and \NameOfMeasurement, derived using varying sample sizes of 4, 8, 16.}
        \label{fig:cos-sim-batch-size}
        \vspace{2ex}
        \includegraphics[width=0.8\textwidth]{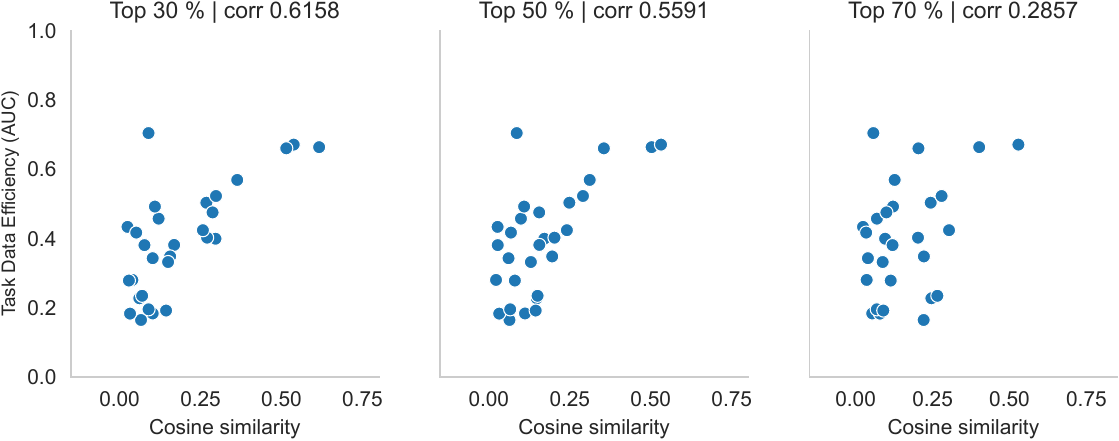}
        \caption{Task data efficiency (AUC) and \NameOfMeasurement, derived from the top 30, 50, 70\% low-confidence segments.}
        \label{fig:cos-sim-conf-seg}
    \end{subfigure}
    \caption{Relationship between the task data efficiency and \NameOfMeasurement, across varying sample sizes and low-confidence segment. The strength of the relationship is captured using Spearman's rank correlation.}
    \label{fig:ablation-auc-cos-correlation}
\end{figure}

\begin{figure}[h]
    \centering
    \begin{subfigure}[t]{0.45\textwidth}
        \centering
        \begin{tabular}{c c c}
        \toprule
        Batch Size &  Mean Abs. Error & p-val\\
        \midrule
        4             &    0.121 & 0.068\\
        8             &    0.111 &  0.015\\
        16            &    0.103 & 0.0048\\
        \bottomrule
        \end{tabular}
        \caption{AUC prediction error by batch size.}
        \label{tab:error-by-batch-size}
        
    \end{subfigure}%
    \hspace{0.05\textwidth}%
    \begin{subfigure}[t]{0.45\textwidth}
        \centering
        \begin{tabular}{c c c}
        \toprule
        Conf. segment &  Mean Abs. Error & p-val\\
        \midrule
        30\%             &    0.095 &  0.00066\\
        50\%            &    0.107 & 0.0064\\
        70\%  &    0.128 & 0.18\\
        \bottomrule
        \end{tabular}
        \caption{AUC prediction error by low-confidence segment threshold.}
        \label{tab:error-by-conf}
    \end{subfigure}
    \caption{The mean absolute AUC prediction error across all downstream tasks using \NameOfMeasurement with varying batch sizes and low-confidence segment thresholds. \cref{tab:error-by-batch-size} and \cref{tab:error-by-conf} display statistically significant predictions (p-value $<$ 0.05) can be made with the sample size as small as 8 and low-confidence threshold as high as 50\%.}
    \label{fig:ablation-auc-pred-error}
\end{figure}

\cref{fig:ablation-auc-cos-correlation} and \cref{fig:ablation-auc-pred-error} demonstrate that \NameOfMeasurement is reasonably robust to both the sample size and the threshold of low-confidence segments. \cref{fig:cos-sim-batch-size} shows that \NameOfMeasurement continues to exhibit a non-random relationship with task data efficiency even when the number of task examples used to compute \NameOfMeasurement is less than 32 (our default). In particular, the prediction of task data efficiency made using as few as 8 or 16 examples remains statistically significant (p-value $<$ 0.05).

Similarly, \NameOfMeasurement derived using confidence thresholds higher than the default top 10\% is predictive of the task data efficiency, as reported in \cref{tab:error-by-conf}. Although the strength of the relationship becomes weaker, using samples from top 50\% low-confidence segment still yields statistically significant meaningful predictions (\cref{tab:error-by-conf}).

\subsection{Calculating \texttt{grad\_norm\_k} and \texttt{conf\_avg\_k} on Low Confidence Examples}
\label{appendix:ablation-low-confidence-segment}

We run additional ablation studies to calculate $\texttt{grad\_norm}_k$ and $\texttt{conf\_avg}_k$;  gradient norm and average model confidence on the low-confidence examples used for \NameOfMeasurement (\cref{tab:overall-AUC-pred-accuracy-by-method-ablation}). Among all task difficulty metrics considered, \NameOfMeasurement (ours) is the most predictive of the data efficiency, which indicates that its predictive power not only comes from the low confidence examples but also from gradient signal conflict from cosine similarity metric.

\begin{wraptable}{r}{7.4cm}
    \centering
        \begin{tabular}{c | c }
        \toprule
        \small
        Methods & Correlation with data efficiency \\
        \midrule
        $\texttt{grad\_norm}_k$             &   -0.1054 $\pm$ 0.041 \\
        $\texttt{conf\_avg}_k$              &   -0.1374 $\pm$ 0.023 \\
        \NameOfMeasurement    &  \textbf{0.675 $\pm$ 0.056} \\
        \bottomrule
        \addlinespace
        \end{tabular}
    \caption{Correlation between data efficiency and task difficulty metrics computed on low-confidence examples.}
    \label{tab:overall-AUC-pred-accuracy-by-method-ablation}
\end{wraptable}

\subsection{Full vs. Low Dimension Model Gradient}
\label{appendix:ablation-gradient-projection}
We explore the amount of information retained or lost due to using rank 64 LoRA gradient or dimensionality reduction technique such as Gaussian random projection, instead of full model gradients when computing \NameOfMeasurement. For Gaussian random projection, we randomly project each layer’s full gradient to a lower dimension and concatenate them into a single vector to compute the metric.

We do not observe a clear advantage in using a much lower dimensional gradient (\cref{tab:gradient-projection-method}), supporting that using low-rank gradients is an effective and efficient way of computing data efficiency predictor, \NameOfMeasurement.

\begin{table}
    \centering
        \begin{tabular}{c | c c}
        \toprule
        \small
        Methods & Correlation with data efficiency & Size of Gradient Vector\\
        \midrule
        Full gradient \texttt{grad\_norm\_k}     &  0.160	 $\pm$ 0.021 & 8GB\\
        Full gradient \NameOfMeasurement         &   0.628 $\pm$ -0.052 & 8GB\\
        Rank 64 \NameOfMeasurement (our choice)  &   0.675 $\pm$ 0.056  & approx. 160M\\
        Random Projection \NameOfMeasurement     &  0.630 $\pm$ 0.053 & approx. 1.6M\\
        \bottomrule
        \addlinespace
        \end{tabular}
    \caption{\NameOfMeasurement computed with rank 64 LoRA gradient outperforms alternate approaches and is relatively memory efficient compared to \NameOfMeasurement computed with full gradient vectors.}
    \label{tab:gradient-projection-method}
\end{table}

\subsection{Increasing the Maximum Data Budget to 10,000}
\label{appendix:ablation-data-budget-10k}

We measure the average raw accuracy at each fine-tuning data size across the 30 tasks, and use tasks with more than 10,000 available data points (15 out of 30) to rerun the experiments end-to-end with the maximum data budget set to 10,000. \cref{fig:max-budget-cos-low} shows the  

\begin{figure}[h]
    \centering
    \begin{subfigure}[t]{1\textwidth}
        \centering
        \begin{tabular}{l|cccccccc}
        \toprule
        Data budget & 50 & 100 & 200 & 500 & 1000 & 2500 & 5000 & 10000 \\
        \midrule
        Avg.\ acc. & 0.155 & 0.180 & 0.233 & 0.322 & 0.360 & 0.400 & 0.412 & 0.415 \\
        \bottomrule
        \addlinespace
        \end{tabular}
        \caption{Avg. raw accuracy at each data budget across the 30 downstream tasks.}
        \label{tab:raw-acc-gain}
    \end{subfigure}%
    \vspace{1ex}
    \begin{subfigure}[t]{1\textwidth}
        \centering
        \begin{tabular}{c|c|c}
        \toprule
        Max. data budget & Corr. with task data efficiency & Abs. mean error \\
        \midrule
        5k budget (30 tasks)  & 0.675 $\pm$ 0.05  & 0.086 $\pm$ 0.030 \\
        10k budget (15 tasks) & 0.699 $\pm$ 0.10  & 0.085 $\pm$ 0.031 \\
        \bottomrule
        \end{tabular}
        \caption{Spearman's rank correlation between \NameOfMeasurement and task AUC, and the AUC prediction error when data efficiency curves are measured with 10,000 as maximum data budget, instead of 5,000.}
        \label{tab:10k-data-budget}
    \end{subfigure}
    \caption{Across the 30 tasks, the accuracy gain is marginal when using more than 5000 fine-tuning data points, and \NameOfMeasurement's high correlation with task AUC and its AUC prediction error remain stable when higher data budget is used (\cref{tab:10k-data-budget}).}
    \label{fig:max-budget-cos-low}
\end{figure}

\subsection{Data efficiency curves under different random seed}

We use a fixed seed (seed=123) when sampling data points to fine-tune for fine-tuning across varying data sizes and use these runs to plot the task data efficiency curves (see \cref{fig:overall-task-data-efficiency}). To account for performance variance due to randomness in sampling, we repeat Llama 3.1 8B Instruct fine-tuning with different random seeds for all data budgets (50, 100, 200, 500, 1000, 2500, 5000) on the 30 downstream tasks. We report the median raw accuracy for each of the three random runs, along with the accuracy differences between the original and the new random seed runs (\cref{tab:random-seed-finetune-runs}). The resulting median differences in task AUCs are negligible, with -0.006 (seed 48) and -0.017 (seed 37) relative to the original task AUCs.

\begin{table}[h]
    \centering
    \begin{tabular}{cccccc}
        \toprule
        \multirow{1}{*}{} & \multicolumn{3}{c}{Median raw accuracy} & \multicolumn{2}{c}{Median diff. in raw accuracy}\\
            
        Data budget & Seed 123 (org) & Seed 48 & Seed 37 & Seed 123 vs. 48  & Seed 123 vs. 37 \\
        \midrule
        0 & 0.492 & 0.619 & 0.558 & 0.026 & 0.026 \\
        50 & 0.67 & 0.728 & 0.708 & 0.028 & 0.028 \\
        100 & 0.702 & 0.73 & 0.747 & 0.024 & 0.032 \\
        200 & 0.784 & 0.777 & 0.771 & 0.023 & 0.020 \\
        500 & 0.816 & 0.818 & 0.817 & 0.028 & 0.018 \\
        1000 & 0.872 & 0.873 & 0.889 & 0.016 & 0.014 \\
        2500 & 0.923 & 0.916 & 0.918 & 0.014 & 0.011 \\
        5000 & 0.937 & 0.930 & 0.927 & 0.010 & 0.009 \\
        \bottomrule
    \end{tabular}
    \caption{Raw accuracy at each data budget across three fine-tuning runs with different random seeds, and the corresponding accuracy difference between the original run (seed=123) and the two additional runs (seed=48, 37).}
    \label{tab:random-seed-finetune-runs}
\end{table}


\section{Generalizing Across Model Families}
\label{appendix:method-to-model-family}

\begin{figure}[h]
    \centering
    \begin{subfigure}{1\textwidth}
        \centering
        \includegraphics[width=0.8\linewidth]{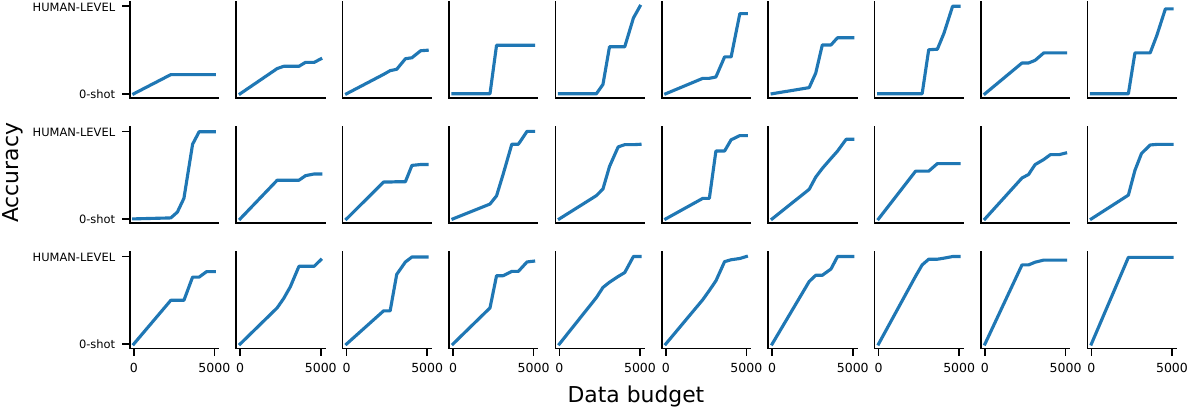}
        \caption{Data Efficiency curves of Mistral 7B Instruct v0.3.}
        \label{fig:mistral-auc-by-metric}
        \vspace{2ex}
        \includegraphics[width=0.8\textwidth]{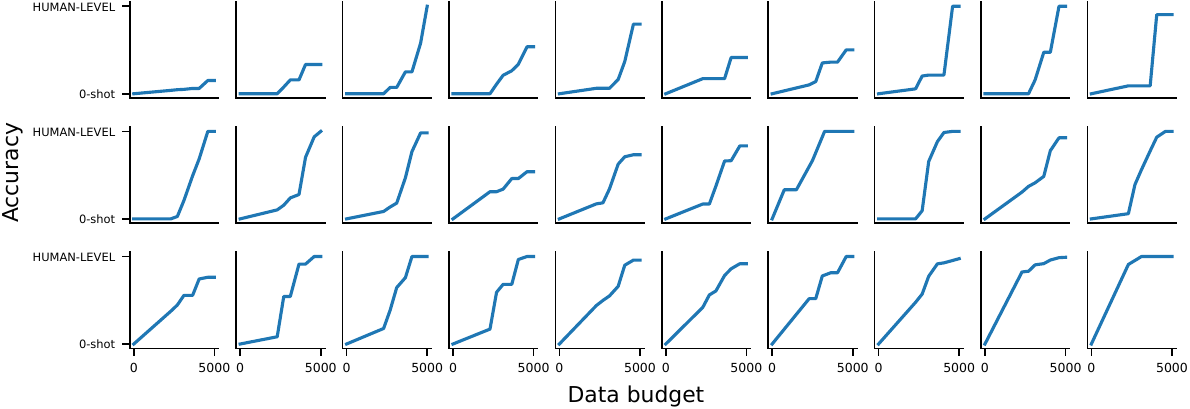}
        \caption{Data Efficiency curves of Qwen 2.5 14B Instruct.}
        \label{fig:qwen-auc-by-metric}
    \end{subfigure}
    \caption{Comparing data budget (from 0 to 5000 examples on log-scale, x-axis) and task performance (from zero-shot to human-level performance, y-axis) across the 30 downstream tasks, for Mistral 7B Instruct v0.3 and Qwen 2.5 14B Instruct.}
    \label{fig:model-families-data-efficiency}
\end{figure}

To assess generalizability across model families, we extend our experiments to the Mistral 7B Instruct v0.3 and Qwen 2.5 14B Instruct. We measure task data efficiency (\cref{fig:model-families-data-efficiency}) and compute corresponding task difficulty metrics to predict data efficiency, using the same compute resources and fine-tuning hyperparameters as in the Llama 3.1 8B Instruct experiments.

\begin{figure}[t]
    \centering
    \begin{subfigure}{1\textwidth}
        \centering
        \includegraphics[width=0.8\linewidth]{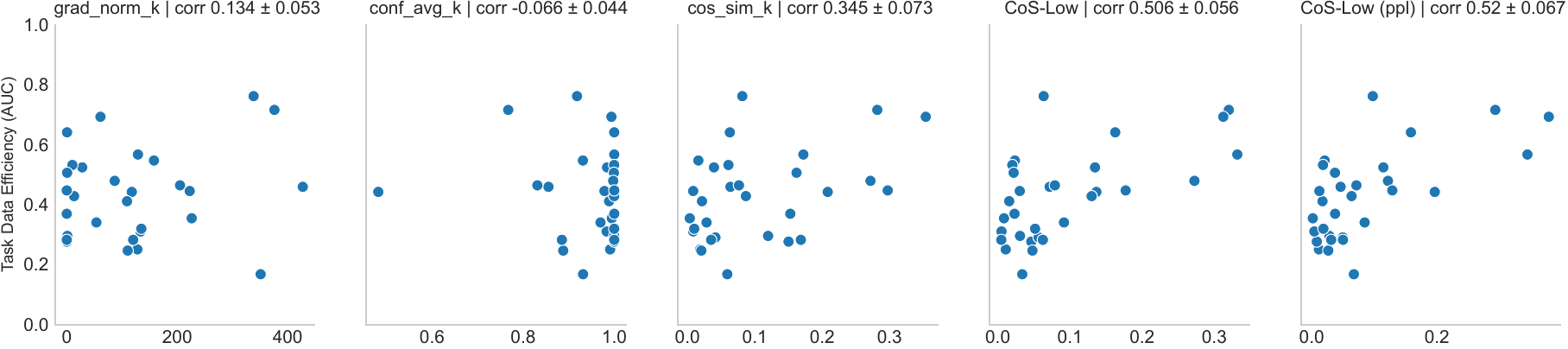}
        \caption{Correlation between task difficulty metrics with data efficiency for Qwen 2.5 14B Instruct.}
        \label{fig:mistral-auc-by-metric}
        \vspace{2ex}
        \includegraphics[width=0.8\textwidth]{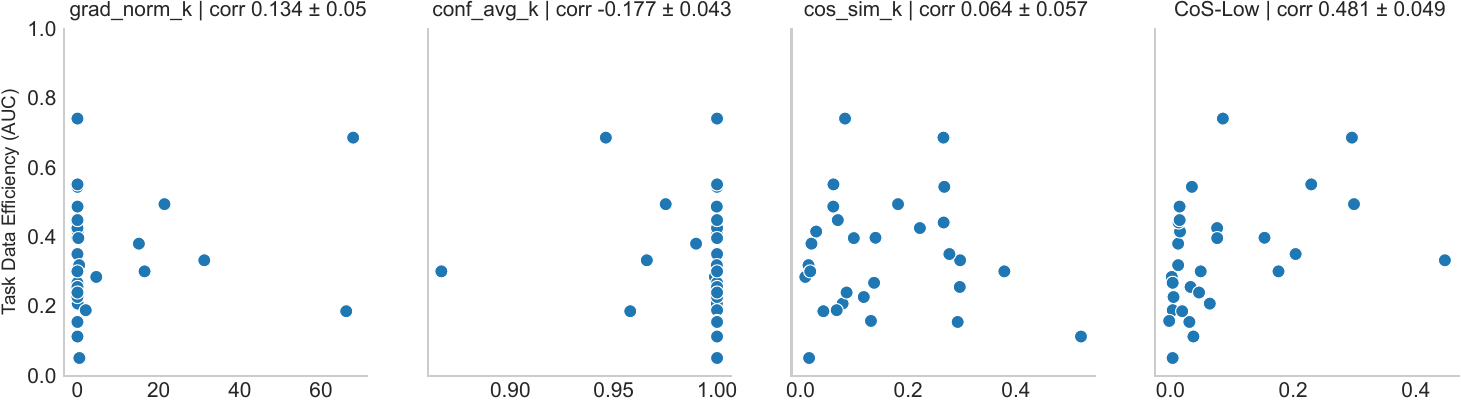}
        \caption{Correlation between tsak difficulty metrics with data efficiency for Mistral 7B Instruct v0.3}
        \label{fig:qwen-auc-by-metric}
    \end{subfigure}
    \caption{\NameOfMeasurement shows the strongest correlation with task data efficiency. The relationship is the strongest, however, for Llama 3.1 8B Instruct, followed by Mistral 7B Instruct v0.3 and Qwen 2.5 14B Instruct.}
    \label{fig:model-families-correlation}
\end{figure}

As shown in \cref{fig:model-families-correlation}, \NameOfMeasurement consistently demonstrates the strongest correlation with task data efficiency and outperforms alternative metrics such as \texttt{grad\_norm\_k}, \texttt{conf\_avg\_k}, and \texttt{cos\_sim\_k}.

While these results demonstrate that task data efficiency prediction using \NameOfMeasurement generalizes beyond the Llama 3.1 8B Instruct model, the relationship between task data efficiency and \NameOfMeasurement is weaker in comparison. One possible explanation is the larger gap between the fine-tuned Mistral and Qwen model performance and human expert-level accuracy for some tasks, due to fine-tuning not improving the performance further from their zero-shot performance.

We hypothesize that the weaker relationship may also be partly attributed to the sensitivity of low-confidence example selection to model-specific tokenization. Our current approach selects low-confidence examples based on the lowest average token probabilities. However, for multi-token outputs, simple averaging does not distinguish between uncertain predictions across all tokens and cases where a single high- or low-confidence token skew the average. The Mistral tokenizer encounters this problem, as it represents multi-digit numbers using multiple tokens. To address this problem, we test perplexity-based confidence estimation to select the low-confidence examples, which provides length-normalized uncertainty estimation. As shown in \cref{fig:mistral-auc-by-metric}, we find that perplexity-based low-confidence example sampling (``\NameOfMeasurement (PPL)'', correlation = 0.52) achieves higher correlation with data data efficiency compared to probability averaging approach (``\NameOfMeasurement'', correlation = 0.5). This improvement suggests that \NameOfMeasurement can benefit from refined low-confidence estimation, especially for tasks involving longer target outputs.


\section{Generalizing to Out-Of-Distribution Tasks}
\label{appendix:method-to-task-categories}

OOD-tasks used to test the generalizability of \NameOfMeasurement beyond the 30 original downstream tasks (\cref{sec:discussion}) consist of two multi-task datasets, four generation tasks, and four domain specific tasks. For all OOD-tasks, we use the model's maximum performance within the 5,000 data budget as the maximum attainable performance proxy.

Each of the \textit{two multitask datasets }consist of five tasks sampled from the 30 downtream tasks, one consisting of  MNIST\_ascii, Boolean\_expressions, Object\_counting, Sports\_understanding, Hyperbaton, and the other of Web\_of\_lies, Reasoning\_about\_colored\_objects, Temporal\_sequences, Tracking\_shuffled\_objects, Formal\_fallacies\_syllogisms\_negation. 

The \textit{four generation tasks} are SQuAD 2.0 \citep{rajpurkar2018knowdontknowunanswerable}, Disfl-QA \citep{gupta2021disflqabenchmarkdatasetunderstanding}, QA WikiData \citep{srivastava2023imitationgamequantifyingextrapolating}, and CoQA \citep{reddy2019coqaconversationalquestionanswering}. For generation task, model perplexity (PPL) on the generated tokens to identify low-confidence segment instead of average probabilities of the generated tokens (see \cref{appendix:estimation-of-model-confidence} for metrics considered to estimate model confidence).

The \textit{four domain-specific tasks} are intent-classification dataset \citep{larson-etal-2019-evaluation}, disaster message categorization \citep{munro12dissertation}, twitter sentiment analysis from HuggingFace (zeroshot/twitter-financial-news-sentiment), and MMLU-Pro \citep{wang2024mmluprorobustchallengingmultitask}. The four downstream tasks are either classification or multiple-choice QnA style tasks, similar to the 30 downstream tasks considered, but do not have a reported human-level performance.

\section{Computation and memory requirement for task difficulty metric calculation}

We assuming backward pass takes twice as much time as forward pass. The main memory requirement is that the model fits in a GPU to be able to run forward and backward pass. The metrics requiring gradients use per-sample gradients of rank 64 LoRAs, takes 1\% of the model weights, and adds only a minor memory requirement. \NameOfMeasurement requires additional forward passes to identify low confidence examples, but the same number of backward pass as it only uses 32 annotated examples for actual metric calculation.

For an 8B model, peak GPU memory storing the rank 64 LoRA gradients of the 32 samples is (BFloat16 memory) * (model parameter requiring gradients) * (batch size) $\approx$ (2) * (8B parameters * 0.02) * (32) $\approx$ 10 GB. The model loaded on GPU adds an extra (BFloat16 memory) * (full model parameter) = 2 * 8 $\approx$ 16GB.

\begin{table}
    \centering
        \begin{tabular}{c | c c | c }
        \toprule
        \small
        \multirow{2}{*}{Method} & \multicolumn{2}{c}{Compute cost} & \multirow{2}{*}{Memory requirement} \\
         & Forward pass &  Backward pass & \\
        \midrule
        $\texttt{grad\_norm}_k$ &  32 * C & 32 * 2C & O(M)\\
        $\texttt{conf\_avg}_k$  &  32 * C & & O(M)\\
        $\texttt{cos\_sim}_k$   &  32 * C & 32 * 2C & O(M)\\
        \NameOfMeasurement      &  2500 * C + 32 * C & 32 * 2C & O(M)\\
        \bottomrule
        \addlinespace
        \end{tabular}
    \caption{Compute and memory requirement of calculating task difficulty metrics. C denotes the time of single forward pass, and M the size of the full model.}
    \label{tab:method-computation-overhead}
\end{table}

\section{Estimating model confidence}
\label{appendix:estimation-of-model-confidence}

In our exploration of model confidence estimation, many alternatives were considered, including model perplexity on its own generation (PPL) and variational ratio for original prediction (VRO). Among these, \NameOfMeasurement on the highest PPL segment showed the strongest correlation with data efficiency than VRO or average softmax probability (our approach). We choose average softmax-probability as confidence proxy for the ease of implementation and reasonably strong correlation with data efficiency; also, it does not require multiple model response generations and requires the least amount of compute. In our experiments, computing PPL required roughly 2x more forward passes, VRO up to 8x.
Nonetheless, PPL may be preferred for tasks involving multi-token outputs, especially as average softmax-probability can be skewed by high or low probability tokens as the generation length increases.

\section{Using \NameOfMeasurement to Concretely Estimate Fine-tuning Data Size}
\label{appendix:cos-low-algorithm}

We describe a high-level algorithm using \NameOfMeasurement and the regression weights learned from ground-truth AUCs across fine-tuning tasks (i.e. \cref{tab:regression-coefficient-across-models}) to predict the concrete fine-tuning data size required to reach the target performance.

\begin{algorithm}
\caption{Predicting Fine-tuning Data Requirements from \NameOfMeasurement}
\begin{algorithmic}[1]
\Require CoS-Low $\ell$, target performance $y$, coefficients $(c, I)$, max budget $N_{\max}$
\Statex
\State \textbf{Step 1: Predict AUC from \NameOfMeasurement}
\State Predicted AUC: $\widehat{\mathrm{AUC}} \gets c \cdot \ell + I$
\Statex
\State \textbf{Step 2: Estimate required data size}
\State \% from data budget needed: $p \gets y^{\frac{\widehat{\mathrm{AUC}}}{1 - \widehat{\mathrm{AUC}}}}$
\State Estimated fine-tuning data size: $n_{\text{required}} \gets 2^{\,p \cdot \log(N_{\max})}$
\State \Return $n_{\text{required}}$
\end{algorithmic}
\end{algorithm}

Note that the log transformation is necessary because the x-axis (fine-tuning data budget) of the performance efficiency curve is on a log-scale (\cref{sec:experiment-setup}). Log scale captures the rapidly changing model performance improvements at smaller data sizes.


\end{document}